\documentclass[runningheads]{llncs}
\usepackage{graphicx}
\usepackage{amssymb}
\usepackage{adjustbox}
\usepackage{float}
\usepackage{color,soul}
\usepackage{todonotes}
\usepackage{listings}
\restylefloat{table}
\usepackage{xcolor}
\usepackage{hyperref}

\begin{document}
\title{STARdom: an architecture for trusted and secure human-centered manufacturing systems}

\titlerunning{STARdom: architecture for secure human-centered manufacturing systems}
%
\author{Jo\v{z}e M. Ro\v{z}anec\inst{1,2,3}\orcidID{0000-0002-3665-639X} \and Patrik Zajec\inst{1,2}
Klemen Kenda\inst{1,2,3}\orcidID{0000-0002-4918-0650} \and Inna Novalija\inst{2} \and Bla\v{z} Fortuna\inst{2,3} \and Dunja Mladeni\'{c}\inst{2} \and Entso Veliou\inst{5}\orcidID{0000-0001-9730-1720} \and Dimitrios Papamartzivanos\inst{4} \and Thanassis Giannetsos\inst{4} \and Sofia Anna Menesidou\inst{4} \and Rubén Alonso\inst{6} \and Nino Cauli\inst{7} \and Diego Reforgiato Recupero\inst{6,7} \and Dimosthenis Kyriazis\inst{8}\orcidID{0000-0001-7019-7214} \and Georgios Sofianidis\inst{8} \and Spyros Theodoropoulos\inst{8,9} \and John Soldatos\inst{10}}
\authorrunning{Jo\v{z}e M. Ro\v{z}anec et al.}
%
\institute{Jo\v{z}ef Stefan International Postgraduate School, Jamova 39, 1000 Ljubljana, Slovenia,\\
\email{joze.rozanec@ijs.si},
\and
Jo\v{z}ef Stefan Institute, Jamova 39, 1000 Ljubljana, Slovenia
\and 
Qlector d.o.o., Rov\v{s}nikova 7, 1000 Ljubljana, Slovenia
\and
Ubitech Ltd, Digital Security \& Trusted Computing Group, Athens, Greece
\and
Department of Informatics and Computer Engineering, University of West Attica, Agiou Spyridonos Street, 12243, Egaleo, Athens, Greece
\and
R2M Solution Srl, Pavia, Italy
\and
Department of Computer Science, University of Cagliari, Cagliari, Italy
\and
Department of Digital Systems, University of Piraeus, Piraeus, Greece
\and
Department of Electrical and Computer Engineering, National Technical University of Athens, Athens, Greece
\and INTRASOFT International, 19.5 KM Markopoulou Ave., GR 19002 Peania, Greece}
\maketitle              
\begin{abstract}
There is a lack of a single architecture specification that addresses the needs of trusted and secure Artificial Intelligence systems with humans in the loop, such as human-centered manufacturing systems at the core of the evolution towards Industry 5.0. To realize this, we propose an architecture that integrates forecasts, Explainable Artificial Intelligence, supports collecting users' feedback and uses Active Learning and Simulated Reality to enhance forecasts and provide decision-making recommendations. The architecture security is addressed as a general concern. We align the proposed architecture with the Big Data Value Association Reference Architecture Model. We tailor it for the domain of demand forecasting and validate it on a real-world case study. 

\end{abstract}

\section{Introduction}\label{INTRODUCTION}
The increasing digitalization of manufacturing has enabled the transition to the fourth industrial revolution (Industry 4.0). Based on Cyber-Physical Systems (CPS) and technologies such as cloud computing, the Industrial Internet of Things (IIoT), and Artificial Intelligence (AI), Industry 4.0 enables flexible production lines and supports innovative functionalities such as mass customization, predictive maintenance, zero-defect manufacturing, and digital twins\cite{soldatos2019digital}. State-of-the-art AI systems in industrial plants operate in controlled environments. Nevertheless, industrial plants' AI systems must be safe, trusted, and secure, even when operating in dynamic, unstructured, and unpredictable environments. Ensuring these systems' safety and reliability is a key prerequisite for deploying them at scale and fully leveraging AI's benefits in manufacturing\cite{hleg2019high}.

The increasing AI adoption in manufacturing has prompted researchers and digital manufacturing communities to research solutions that boost the development of secure and trusted AI systems in production lines and their compliance to ethical principles. A prominent example is the surge of research on Explainable AI (XAI), a field of AI concerned with building models transparent to users or techniques that provide insights on key factors influencing the model's forecast. Such insights help to assess the soundness of given forecasts and aid human decision-making. In the particular manufacturing case, XAI boosts AI solutions' transparency and increases human workers' acceptance. Though much research was done regarding XAI techniques, there is still little research on how the forecast explanations affect human decision-making. To assess their impact, human feedback is required. Human feedback collection regarding AI models and forecast explanations can be addressed as an Active Learning (AL) problem. AL enables AI systems to operate in the absence of enough labeled data and consult an authoritative source (e.g., query a human expert) to obtain the required annotations. Focusing knowledge acquisition only on the most promising data instances accelerates knowledge acquisition and increases industrial systems' robustness.

The increasing digitalization and the development and deployment of AI models have increased vulnerabilities to cyber-attacks. In particular, AI models are vulnerable to attacks in the training phase (e.g., poisoning) and operational phase (e.g., evasion). AI models and XAI techniques must be robust against adversarial attacks. 

The development, deployment, and operation of efficient industrial systems that combine advanced AI systems with appropriate cyber-defense strategies must be grounded in well-structured architectures. The latter boost the scalability and efficiency of the systems while at the same time facilitating their integration. In recent years, standards-development organizations, industrial associations, and research groups have introduced reference architecture models for industrial systems. For instance, the Reference Architecture Model for Industry 4.0 (RAMI 4.0)\cite{schweichhart2016reference} illustrates the main building blocks of Industry 4.0 systems and presents how to develop an industrial system in a structured way. As another example, the Industrial Internet Consortium (IIC) has introduced the Industrial Internet Reference Architecture (IIRA)\cite{IIRA} that specifies a common architecture framework for developing interoperable IoT systems for different industry verticals, including manufacturing. These architectures do not directly address the security and safety aspects of industrial systems. Complementary architectural frameworks were developed to address them, such as the Industrial Internet Security Framework (IISF)\cite{IISF16}. Likewise, these architectures do not address the structuring principles of AI systems (e.g., the building blocks of machine learning pipelines). These principles are also addressed separately in specifications like the Big Data Value Association (BDVA) Reference Architecture. Overall, there is no single architecture specification that addresses the needs of trusted and secure AI systems with humans in the loop, such as human-centered manufacturing systems at the core of the evolution towards Industry 5.0 \cite{nahavandi2019industry,EC2020}. 

This paper aims to alleviate this gap based on the description and initial validation of an architecture for trusted and human-centered AI systems in production environments. The architecture aims to serve as a blueprint for realizing manufacturing systems that support AI models, XAI, and AL while providing security guarantees. It provides structuring principles for integrating these modules. It creates synergies between them, aligned with some of the above-listed reference models, such as the BDVA reference architecture model and the IISF. Specifically, the introduced architecture leverages layering concepts and components from these architectures towards providing a blueprint for developing trusted AI systems. Leveraging on these blueprints, we also illustrate how cybersecurity components for industrial AI systems are combined with XAI and AL to enable trusted, human-centered AI. The architecture is instantiated and validated on a real-world use case regarding demand forecasting.

The rest of this paper is structured as follows: Section~\ref{RELATED-WORK} presents related work, and Section~\ref{ARCHITECTURE} describes the proposed architecture and its application to demand forecasting. Finally, in Section~\ref{CONCLUSION}, we discuss an implementation, provide our conclusions and outline future work.

\section{Related Work}\label{RELATED-WORK}
The increasing digitalization of manufacturing enables the creation of AI models for multiple use cases. When decisions are made based on forecasts, they should be accompanied with explanations on how those forecasts were reached by the model\cite{ribeiro2016should,lundberg2017unified}. Such explanations can help the user understand the model reasoning, evaluate the soundness of the forecasts and explanation provided, increasing trust in the model, and avoid costly mistakes. Explanations must be contextual to the user\cite{henin2021multi}. They can inform key factors influencing the forecast, emphasize the actionable aspects and provide examples of how changes in certain variables can change the forecast outcome. Forecast explanation's quality is a matter of ongoing research, where user feedback can be crucial. To realize it,\cite{tulli2020learning} proposed a framework with three components: a forecasting engine, an explanation engine, and a feedback loop to learn from the users.

Two core ideas drive research in AL: (i) the learner can benefit from asking questions, and (ii) unlabeled data is available or easily obtained\cite{settles2009active}. AL strategies can selectively choose the items to be presented to the user, for which feedback is expected. Since users are usually reluctant to provide it, AL strategies enable to identify a small set of items on which users' input can convey the system valuable information\cite{elahi2016survey}. In particular, AL can collect feedback for given forecasts, forecast explanations, and decision-making options.

While AL is concerned with selecting the most informative unlabeled data instances and querying labels for them, simulated reality provides the capability to generate synthetic data in a supervised setting. Such data augmentation is valuable, e.g., in the context of imbalanced datasets, such as infrequent demand items or visual quality inspection\cite{yun2020automated}. Simulated reality is also a key component for reinforcement learning (RL). Simulations enable RL agents to explore many episodes of trial and error to explore and learn efficient policies. By envisioning the consequences of an action, simulations can help to validate desired outcomes in a real-world setting\cite{amodei2016concrete}.

The richness and connectivity enabled by the increased manufacturing digitalization and IoT-based \textit{Systems-of-Systems} pose a significant risk regarding malware. Malware can target the network layer and edge devices to extract sensitive information (impacting confidentiality) or alternate data originated by those devices (impacting trustworthiness). AI models are also subject to cyberattacks, such as poisoning and evasion\cite{zhang2019adversarial}. The goal of such attacks is to disrupt the AI model forecasts to mislead dependent systems and decision-making. Poisoning consists of altering the training data to disrupt AI models learning and their later performance. Evasion, on the other side, aims to produce wrong forecasts on already trained models by providing a carefully perturbed input that confuses the AI model.

The development of conversational multimodal interfaces can enhance human-machine interactions\cite{maurtua2017natural}. While an output interface can be provided through different modes, conversational interactions allow obtaining complementary information from the user, who can provide locally observed collective knowledge\cite{preece2015sherlock} not captured by other means.

To realize the components and interactions listed above, an architecture model must be developed, following the best practices introduced by reference architectures described in Section~\ref{INTRODUCTION}. We introduce such an architecture in Section~\ref{ARCHITECTURE}.

\section{Proposed Architecture}\label{ARCHITECTURE}

\begin{figure*}[!t]
\centering
\includegraphics[width=4.2in,height=1.9in]{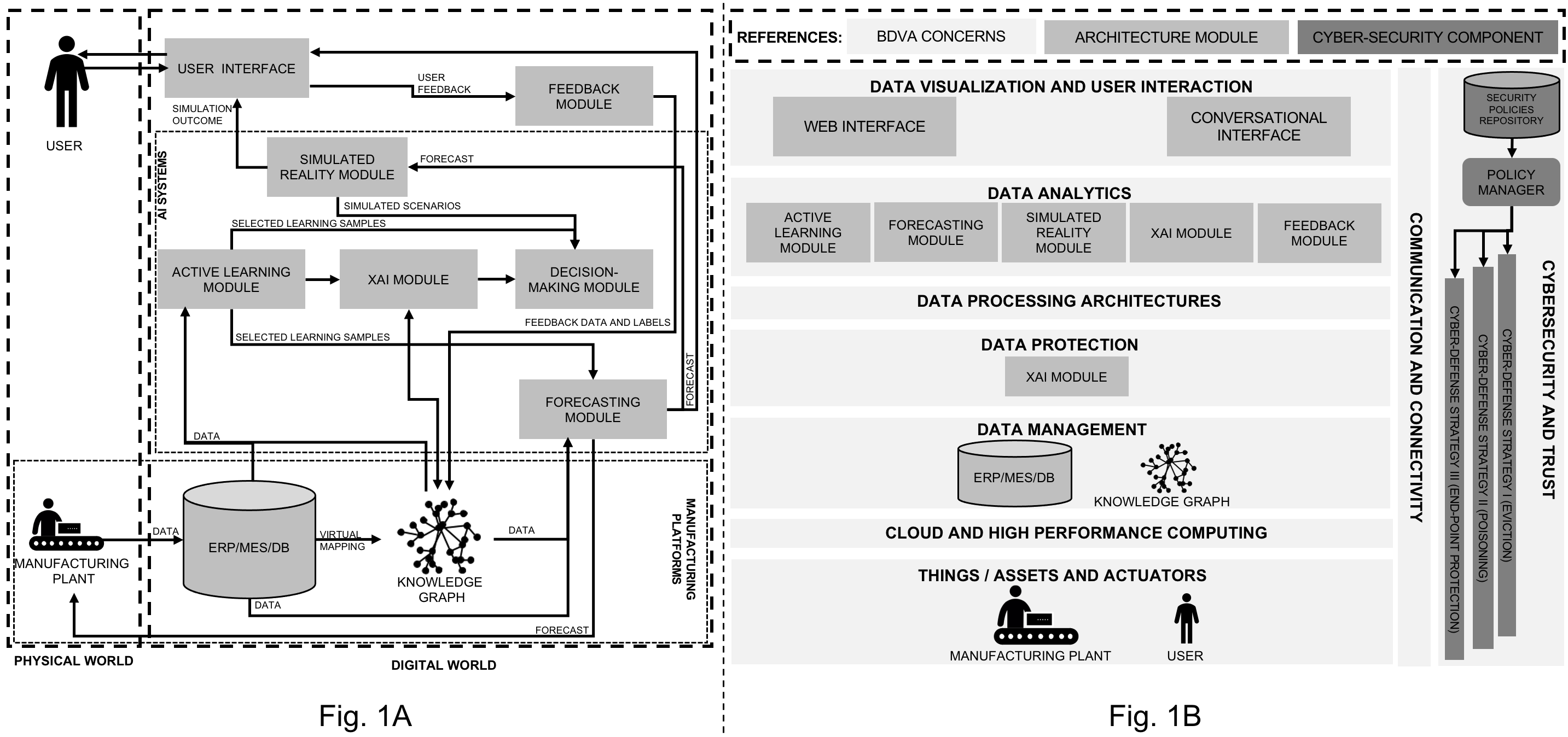}
\caption{Proposed architecture. Fig.~1A displays physical and digital components and their interactions. Fig.~1B shows how components in Fig. 1A relate to the BDVA reference architecture model.}
\label{F:ARCHITECTURE}
\end{figure*}

We propose a modular architecture to enable efficient industrial systems that combine advanced AI systems with appropriate cyber defense strategies. The architecture integrates predictions, gathers insights on relevant features, incorporates domain knowledge and context to each prediction, and provides a forecast explanation and decision-making options to the end-user. Cyber-security is considered across all modules. The architecture (see Fig.~\ref{F:ARCHITECTURE}) comprises the following components:

\begin{itemize}
  \item \textbf{Knowledge Graph}: receives data from Manufacturing Execution Systems (MES), Enterprise Resource Planning (ERP) software, and other systems and from the \textit{Feedback module}, which is mapped into a graph considering relevant concepts and relationships between them. The knowledge graph can be extended for specific use cases, leveraging existing knowledge and enriching it with new domain knowledge and data specific to that case.
  
  \item \textbf{Forecasting Module}: provides forecasts based on AI and simulation models. Different models are developed regarding the task to be solved (classification, regression, clustering, or ranking). The models require past data to learn patterns and predict future outcomes based on them. Inputs can be obtained from the knowledge graph, databases, and external systems, such as ERP and MES.
  
  \item \textbf{XAI Module}: receives input from the \textit{Forecasting module} to assess which features are most relevant to a given forecast. Based on them, an explanation is created for the user, considering their profile, to ensure relevant information is display and confidentiality protected. The XAI module also receives input from the \textit{Feedback module}, through which the users assess the goodness of explanations provided and gather knowledge on actions yet unknown to the system. Such feedback can be used to measure forecast explanation's quality, detect biases in the forecasting model, explainability algorithms, and be used to correct and tailor future explanations and detect potential poisoning and evasion attacks.

  \textbf{Decision-making module}: envisioned as a recommender system. The decision-making module considers inputs from the \textit{Forecasting}, \textit{XAI}, \textit{Simulated reality}, and \textit{Feedback} modules, to provide decision-making advice to users based on their profile and context. Based on inputs from the \textit{XAI module} can create directive explanations\cite{singh2021directive}, suggesting decision-making options that lead to a desired outcome.
  
  \item \textbf{Simulated Reality Module}: receives input from the \textit{Forecasting module}, which could be either a set of predictions or candidate actions. This input is used to generate alternative scenarios, either through the generation of synthetic data or the reconfiguration of an RL agent environment. These scenarios are evaluated, and the confidence of the algorithm's predictions is assessed. The human operator is an active participant in this process, verifying the simulated scenarios' plausibility and monitoring the algorithm's reactions to them. The output of the component consists of plausible, novel, or low confidence scenarios provided to the \textit{Decision-making module} as a means of robustifying the algorithm's predictive capacity and covering its blind spots before an action is taken in the real world. The algorithm can use the simulation component, either as a means of data augmentation or for readjusting to novel or anomalous inputs.
  
  \item \textbf{Active Learning Module}: monitors existing knowledge and labeled data to suggest which data instances will be displayed to users to get their feedback and labels. It prioritizes unlabeled data for data instances expected to contain interesting information. Collected feedback and new labeled instances are consumed by the \textit{Forecasting}, \textit{XAI}, and \textit{Decision-making} modules.
  
  \item \textbf{Feedback Module}: collects feedback from the users regarding forecasts, forecast explanations, and the decision-making options provided. Feedback can be explicit or implicit, and can be used by the \textit{Forecasting}, \textit{Active Learning}, \textit{XAI} and \textit{Decision-making} modules to enhance their operation. 
  
   \item \textbf{User interface}: enables users' multimodal interactions with the system, e.g., interact through the use of their voice, complemented by other modalities, such as on-screen forms. It enables the machine to provide information to the user through audio, natural language, or other means such as visual information.
\end{itemize}

The IISF provides a guide for understanding and implementing security for systems that comply with the IIRA. In particular, the IISF provides guidelines for securing each component of the IIRA while at the same time binding these components together in a trustworthy system. They emphasize means to secure the traditional Operation Technology (OT), as conventional IT security solutions do not apply directly to OT systems and services.
The IISF specifies a range of functionalities applied across all industry components in a horizontal approach, i.e., a cross-cutting function. These functionalities include protecting data, protecting (edge/cloud) endpoints, protecting communications and connectivity, security configuration, management, monitoring and analytics.
Our proposed architecture aligns with IISF. It defines a cyber-defense layer destined to protect AI systems from cybersecurity attacks such as poisoning and evasion attacks.

As depicted in Fig.~\ref{F:ARCHITECTURE}B, the proposed architecture is well aligned with the BDVA reference architecture model and thus can be directly utilized in various additional domains, even beyond the Industry 4.0 domain. The BDVA architecture exploits as a backbone Cloud, High-Performance Computing environments, and data sources and devices in the IoT and edge spaces. These will be exploited in the proposed architecture's scope to deploy the corresponding modules and the diverse data stores – including the Knowledge Graph and the data processing that needs to be performed on top of them. Aligning on that level of the BDVA reference models facilitates the direct deployment of the architecture modules  described above and presented in Fig.~\ref{F:ARCHITECTURE}A. Of major importance are the components on the data analytics layer of the BDVA reference model, also addressing the \textit{BDVA Reference Model Verticals} in terms of time series, IoT, media, and other datasets. Thus, the modules placed on the data analytics layer of BDVA tackle the various types of data and can be utilized in combination with analytics models and techniques in various domains to provide additional information (e.g., explainability outcomes). Additionally, the proposed architecture enhances the BDVA reference model in Data Visualization and User Interaction by delivering a conversational interface to obtain feedback from the users and utilize it in the data analytics scope.

We partially validated this architecture with an implementation based on EU HORIZON 2020 FACTLOG and STAR partners' real-world data. The implementation comprised demand forecasting models providing daily forecasts. We created forecast explanations based on LIME\cite{ribeiro2016should} implemented decision-making recommendations regarding manufactured goods transportation\cite{rozanecACM}. Finally, we implemented a knowledge graph to link existing data semantically and an interface to collect implicit and explicit feedback regarding forecasts, forecast explanations, and decision-making options\footnote{A video of the application was published in \url{https://youtu.be/ysD2oXQO98I}}. While a productive version of the demand forecasting models exists, the latest forecasting models and the remaining components were not deployed into production environments. AL and simulated reality modules were not implemented.

\section{Conclusions}\label{CONCLUSION}
This research introduces an architecture that integrates different components to enable trusted and human-centered AI. In particular, we align the architecture with the BDVA Reference Model to be directly utilized in additional domains beyond the Industry 4.0 domain. Most architecture components were instantiated and validated on a real-world use case. Future work will focus on developing AL and simulated reality modules and cyber-security enhancements. This architecture will be applied to multiple use cases, such as defect detection towards the Quality 4.0 paradigm, in the EU H2020 STAR project.

\section*{Acknowledgement}
This work was supported by the Slovenian Research Agency and the European Union’s Horizon 2020 program projects FACTLOG and STAR under grant agreements numbers H2020-869951 and H2020-956573.

%
%
\bibliographystyle{splncs04}
\bibliography{main}

\begin{thebibliography}{10}
\providecommand{\url}[1]{\texttt{#1}}
\providecommand{\urlprefix}{URL }
\providecommand{\doi}[1]{https://doi.org/#1}

\bibitem{EC2020}
European commission, enabling technologies for industry 5.0, results of a
  workshop with europe's technology leaders.
  \url{https://op.europa.eu/en/publication-detail/-/publication/8e5de100-2a1c-11eb-9d7e-01aa75ed71a1/language-en},
  september 2020

\bibitem{IIRA}
The industrial internet of things volume g1: Reference architecture.
  \url{https://www.iiconsortium.org/pdf/IIRA-v1.9.pdf}, version 1.9 June 19,
  2019

\bibitem{IISF16}
Industrial internet of things, volume g4: Security framework, industrial
  internet consortium. IIC:PUB:G4:V1.0:PB:20160926, september 2016

\bibitem{amodei2016concrete}
Amodei, D., Olah, C., Steinhardt, J., Christiano, P., Schulman, J., Man{\'e},
  D.: Concrete problems in ai safety. arXiv preprint arXiv:1606.06565  (2016)

\bibitem{elahi2016survey}
Elahi, M., Ricci, F., Rubens, N.: A survey of active learning in collaborative
  filtering recommender systems. Computer Science Review  \textbf{20},  29--50
  (2016)

\bibitem{henin2021multi}
Henin, C., Le~M{\'e}tayer, D.: A multi-layered approach for tailored black-box
  explanations  (2021)

\bibitem{hleg2019high}
HLEG, A.: High-level expert group on artificial intelligence: Ethics guidelines
  for trustworthy ai. European Commission, 09.04  (2019)

\bibitem{lundberg2017unified}
Lundberg, S., Lee, S.I.: A unified approach to interpreting model predictions.
  arXiv preprint arXiv:1705.07874  (2017)

\bibitem{maurtua2017natural}
Maurtua, I., Fernandez, I., Tellaeche, A., Kildal, J., Susperregi, L.,
  Ibarguren, A., Sierra, B.: Natural multimodal communication for human--robot
  collaboration. International Journal of Advanced Robotic Systems
  \textbf{14}(4),  1729881417716043 (2017)

\bibitem{nahavandi2019industry}
Nahavandi, S.: Industry 5.0—a human-centric solution. Sustainability
  \textbf{11}(16), ~4371 (2019)

\bibitem{preece2015sherlock}
Preece, A., Webberley, W., Braines, D., Hu, N., La~Porta, T., Zaroukian, E.,
  Bakdash, J.: Sherlock: Simple human experiments regarding locally observed
  collective knowledge. Tech. rep., US Army Research Laboratory Aberdeen
  Proving Ground, United States (2015)

\bibitem{ribeiro2016should}
Ribeiro, M.T., Singh, S., Guestrin, C.: " why should i trust you?" explaining
  the predictions of any classifier. In: Proceedings of the 22nd ACM SIGKDD
  international conference on knowledge discovery and data mining. pp.
  1135--1144 (2016)

\bibitem{rozanecACM}
Ro{\v{z}}anec, J.: Explainable demand forecasting: A data mining goldmine.
  Companion Proceedings of the Web Conference 2021 (WWW '21 Companion), April
  19--23, 2021, Ljubljana, Slovenia  (2021). \doi{10.1145/3442442.3453708}

\bibitem{schweichhart2016reference}
Schweichhart, K.: Reference architectural model industrie 4.0 (rami 4.0). An
  Introduction. Available online: https://www. plattform-i40. de I  \textbf{40}
  (2016)

\bibitem{settles2009active}
Settles, B.: Active learning literature survey  (2009)

\bibitem{singh2021directive}
Singh, R., Dourish, P., Howe, P., Miller, T., Sonenberg, L., Velloso, E.,
  Vetere, F.: Directive explanations for actionable explainability in machine
  learning applications. arXiv preprint arXiv:2102.02671  (2021)

\bibitem{soldatos2019digital}
Soldatos, J., Lazaro, O., Cavadini, F., Boschi, F., Taisch, M., Fantini, P.M.,
  et~al.: The Digital Shopfloor: Industrial Automation in the Industry 4.0 Era.
  Performance Analysis and Applications. River Publishers Series in Automation,
  Control and Robotics (2019)

\bibitem{tulli2020learning}
Tulli, S., Wallk{\"o}tter, S., Paiva, A., Melo, F.S., Chetouani, M.: Learning
  from explanations and demonstrations: A pilot study. In: 2nd Workshop on
  Interactive Natural Language Technology for Explainable Artificial
  Intelligence. pp. 61--66 (2020)

\bibitem{yun2020automated}
Yun, J.P., Shin, W.C., Koo, G., Kim, M.S., Lee, C., Lee, S.J.: Automated defect
  inspection system for metal surfaces based on deep learning and data
  augmentation. Journal of Manufacturing Systems  \textbf{55},  317--324 (2020)

\bibitem{zhang2019adversarial}
Zhang, J., Li, C.: Adversarial examples: Opportunities and challenges. IEEE
  transactions on neural networks and learning systems  \textbf{31}(7),
  2578--2593 (2019)

\end{thebibliography}
\end{document}